# Simultaneous Facial Landmark Detection, Pose and Deformation Estimation under Facial Occlusion

Yue Wu
ECSE Department
Rensselaer Polytechnic Institute
wuyuesophia@gmail.com

Chao Gou
Institute of Automation
Chinese Academy of Sciences
gouchao2012@ic.ac.cn

Qiang Ji
ECSE Department
Rensselaer Polytechnic Institute
jiq@rpi.edu

## Abstract

*Facial landmark detection, head pose estimation, and facial deformation analysis are typical facial behavior analysis tasks in computer vision. The existing methods usually perform each task independently and sequentially, ignoring their interactions. To tackle this problem, we propose a unified framework for simultaneous facial landmark detection, head pose estimation, and facial deformation analysis, and the proposed model is robust to facial occlusion. Following a cascade procedure augmented with model-based head pose estimation, we iteratively update the facial landmark locations, facial occlusion, head pose and facial deformation until convergence. The experimental results on benchmark databases demonstrate the effectiveness of the proposed method for simultaneous facial landmark detection, head pose and facial deformation estimation, even if the images are under facial occlusion.*

## 1. Introduction

Typical facial behavior analysis tasks include facial landmark detection, head pose estimation, and facial deformation analysis. Facial landmark detection aims to detect the key points around facial components and facial contour. The goal of head pose estimation is to predict the orientation and translation of the head with respect to the camera coordinate frame. Facial deformation refers to the non-rigid facial motion induced by facial expression change. Current research usually tackles these tasks independently and sequentially. In reality, rigid head movements and non-rigid facial expressions often happen together and they both affect the 2D facial landmarks (see Figure 1). Head pose and non-rigid facial deformation can lead to different 2D facial shapes, and the 2D facial landmark locations can reflect the head pose and non-rigid facial deformation. Therefore, due to the coupled interactions, facial landmark, head pose, and facial deformation should be estimated jointly utilizing their joint relationships.

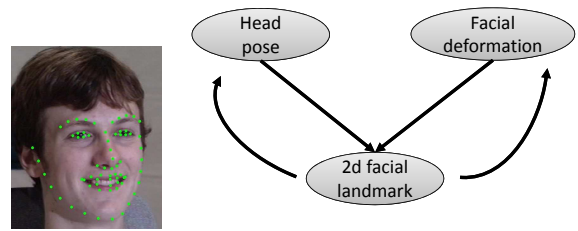

Figure 1. The coupling of facial landmark locations, head pose and facial deformation

Facial occlusion brings significant challenges for facial behavior analysis. As shown in Figure 2, facial occlusion can be induced by objects or it may be self-occlusion due to significant head poses (e.g. > 60 degree). Facial appearance information is noisy on images with occlusion, and facial shape on the occluded facial parts is difficult to estimate. The facial occlusion causes problems for all facial analysis tasks, including facial landmark detection, head pose estimation, and facial deformation analysis.

Based on the intuitions above, we propose to simultaneously estimate the facial landmark locations, head pose and facial deformation in a unified framework using a method that is robust to facial occlusion. There are a few major contributions of the proposed method:

- We propose an iterative cascade method to simultaneously perform facial landmark detection, pose and deformation estimation. The unified framework can leverage the joint relationships among landmarks, pose and deformation to boost the performances of all the tasks. This is in contrast to most of the existing works, which treat them sequentially or independently. It is also different from some joint methods performing one-shot estimation [18][2].

- The proposed method allows us to systematically integrate the learning-based facial landmark detection with model-based head pose and facial deformation estima-tion without the need for 3D annotation. This is in contrast to the existing data-driven learning-based methods that rely on 3D annotations [30][14][22][29].

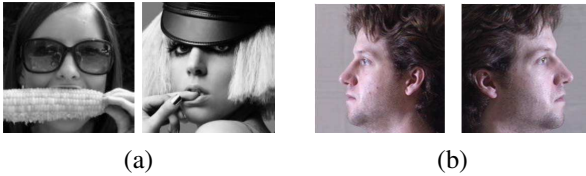

Figure 2. Facial occlusion induced by (a) object occlusion (COFW database [3]) or (b) self-occlusion due to extreme head poses (MultiPIE database [10]).

- Unlike most of the existing works [22][29][18][2], our method explicitly estimates facial occlusion, which can, in turn, help landmark detection, pose and deformation estimation under facial occlusion.

- The experiments on benchmark databases demonstrate the effectiveness of the proposed method for landmark detection, pose estimation, and deformation estimation under facial occlusion.

The remaining part of the paper is organized as follows. In section 2, we review the related works. In section 3, we introduce the proposed method. In section 4, we evaluate the proposed method and compare it to other state-of-the-art works. In section 5, we conclude the paper.

## 2. Related Work

### 2.1. Facial landmark detection

Facial landmark detection algorithms can be classified into the holistic methods [5][23], Constrained Local Methods (CLM) [6][18][2] and regression-based methods [26][17][1][4]. The proposed method follows the regression framework.

Some recent facial landmark detection algorithms try to handle facial occlusion. The method in [3] is one early work that predicts the occlusion labels of different facial landmarks. The authors train nine occlusion-dependent models based on part of the facial appearance, and combine them with weights based on the estimated facial occlusion for landmark detection. In [9], a probabilistic graphical model is utilized to model the spatial relationship among facial landmarks and their occlusions under different facial deformation and expressions for robust landmark detection under facial occlusion. The method in [25] iteratively estimates the facial landmark occlusion and locations following the cascade regression framework. However, this method does not consider head pose and non-rigid facial deformation.

### 2.2. Head pose estimation

Head pose estimation algorithms can be classified into learning-based methods and model-based methods. The learning-based methods utilize pattern recognition and machine learning techniques to directly map the image appearance to the discrete head pose (e.g., left pose, frontal, and right pose) or continuous pose angles (e.g., pitch, yaw, and roll). Common learning techniques include multi-layer perceptron [19], random forests [7] and the Partial Least Squares (PLS) regression method [11] etc.

Model-based methods utilize 2D facial landmarks and 3D computer vision techniques for pose estimation. For example, in [24], the 3D head pose and the facial deformation of a driver are estimated based on a flexible model and detected 2D facial landmarks. In [21], a general 3D face with six landmarks is combined with the RANSAC method for head pose estimation. The existing head pose estimation algorithms usually do not explicitly handle facial occlusion. The model-based approaches may fail on facial images with extreme head poses, since they may not be able to exclude the landmarks on the self-occluded facial parts.

### 2.3. Joint estimation

There are a few algorithms that combine head pose or deformable models with facial landmark detection. They can be roughly classified into 3D Constrained Local Methods (3D CLMs), 3D cascade regression methods, and other methods.

3D CLMs [18][2][28] use the 3D facial shape deformable model in combination with the head pose parameters to regularize the 2D facial landmark locations for 2D facial landmark detection. While traditional CLMs [6] predict the 2D deformable model coefficients for landmark detection, the 3D CLMs can predict both the 3D model coefficients and head pose angles. Even though both the 3D CLMs [18][2][28] and the proposed method can predict head pose and deformable models coefficients, there are a few differences. While the goal of 3D CLMs is facial landmark detection and the head pose estimation is only the intermediate result, the proposed method aims to perform both landmark detection and pose estimation. Pose estimation accuracy is usually not reported in 3D CLMs [18][28]. Furthermore, while 3D CLMs perform one-step estimation for pose and deformable coefficients to directly determine the 2D landmark locations, the proposed method follows the regression approaches and performs cascade regression to gradually update the landmark locations, head poses, and deformable model coefficients. Since 3D CLMs directly predict model coefficients and head pose for landmark detection, and small model coefficients and pose errors may lead to large landmark detection errors, their estimation may not be accurate. On the other hand, our method directly predicts landmarks, which may be more accurate.

3D cascade regression methods are learning-based methods that perform 3D prediction. They either predict the 3D facial landmark locations that determine the head pose and

3472

3D model coefficients, or they predict head pose and deformable coefficients that determine the 3D facial landmark locations. For example, in [22], Tulyakov and Sebe propose to predict the 3D facial shape from the facial image using cascade regression methods based on 3D training data, and the head pose can be calculated based on the predicted 3D facial shape. In [13][30] [14], cascade regression learning methods are proposed to directly predict the pose and 3D model coefficients with 3D training data. There are a few differences between the 3D cascade regression methods and the proposed method. For example, 3D cascade regression methods are purely data-driven, and they require 3D facial landmark annotations or head pose annotations for each training data. Since the proposed method combines learning with model-based head pose estimation, it does not require 3D annotations. Similar to the 3D CLMs, the 3D cascade regression methods mainly focus on facial landmark detection and head pose estimation results may not be reported. It is also not clear how them perform on images with facial occlusion.

There are a few other joint learning methods that try to simultaneously predict 2D facial landmarks and pose from image appearance. For example, in [31][12], 2D head pose-dependent facial landmark detection models are constructed and applied to the testing image. The final selected 2D landmark detection results and head pose are determined by the pose-dependent model with the smallest fitting error. In [29], a cascade iterative framework is proposed to predict 2D landmarks, pose, and expression labels based on random forests. Compared to the proposed method, the methods in this category are purely learning-based approaches requiring additional pose labels, ignoring the projection model, while the proposed method combines learning and model without the need for pose annotations. In addition, while those methods can only predict the discrete head poses, the proposed method can predict the continuous head pose. It is also not clear how the method in [29] will perform on images with occlusion.

## 3. Approach

### 3.1. General framework

The proposed method simultaneously performs facial landmark detection, head pose estimation, and facial deformation estimation under facial occlusion given the facial image, denoted as $I$. The facial landmark locations of the facial key points on 2D images are denoted as $\mathbf{x} \in \mathbf{R}^D$, where $D$ is the number of landmark points. The head pose refers to the continuous pose angles, including pitch, yaw, and roll, denoted as $\mathbf{h} = \{pitch, yaw, roll\}$. The facial deformation refers to the non-rigid facial deformation related to facial expression, facial action unit etc., excluding the rigid pose variations. In particular, assume we have a 3D deformable facial model trained with 3D facial shapes using principal component analysis technique, the 3D face shape can be represented using the deformation coefficients, denoted as $\alpha$.

$$\mathbf{s} = \bar{\mathbf{s}} + B\alpha \tag{1}$$

Here, $\mathbf{s} = \{x_1, y_1, z_1, ..., x_D, y_D, z_D\}^T$ denotes the 3D face shape. $\bar{\mathbf{s}}$ is the average 3D shape and $B$ represents the orthonormal bases. To handle facial occlusion, we introduce the landmark visibility vector $\mathbf{c} \in [0,1]^D$, which specifies the probabilities that the landmark points are visible. The proposed algorithm simultaneously estimates $\mathbf{x}$, $\mathbf{h}$, $\alpha$, and $\mathbf{c}$.

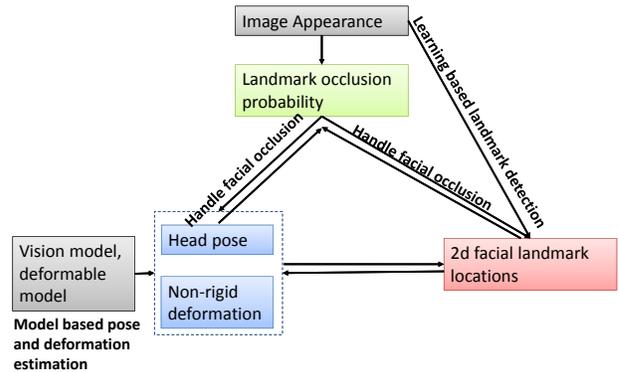

Figure 3. Overall framework

The proposed method follows a cascade iterative procedure. The overall algorithm is illustrated in Algorithm 1 and Figure 3. We first initialize the facial landmark locations using the mean face $\mathbf{x}^0$, assuming the pose is frontal $\mathbf{h}^0 = \mathbf{0}$, there is no non-rigid deformation $\alpha^0 = \mathbf{0}$, and all the landmarks are visible $\mathbf{c}^0 = \mathbf{1}$. Then, we iteratively update the landmark visibility probabilities, the landmark locations, the head pose angles and non-rigid deformations. For each iteration $t$, we sequentially update the landmark visibility probabilities, the landmark locations, and the head poses. In the first step, when updating the landmark visibility probabilities $\mathbf{c}^t$, we predict the landmark visibility probability updates $\Delta \mathbf{c}^t$ given the previously predicted landmark locations $\mathbf{x}^{t-1}$ and the head pose $\mathbf{h}^{t-1}$. The estimated visibility probability updates $\Delta \mathbf{c}^t$ will be added to the previously estimated visibility probabilities $\mathbf{c}^{t-1}$ to generate the new visibility probabilities $\mathbf{c}^t$. In the second step, when updating the landmark locations $\mathbf{x}^t$, we predict the landmark location updates $\Delta \mathbf{x}^t$ by using the previously estimated landmark locations $\mathbf{x}^{t-1}$, the currently estimated landmark visibility probabilities $\mathbf{c}^t$, the previously estimated non-rigid deformation information $\alpha^{t-1}$, and the previously estimated head pose $\mathbf{h}^{t-1}$. The update will be added to the previous estimation to generate the new estimation of landmark locations $\mathbf{x}^t$. In the third step, when updating the head pose $\mathbf{h}^t$ and

3473

**Algorithm 1:** The general framework

1. Initialize the landmark locations $\mathbf{x}^0$ using the mean face; Assuming all the landmarks are visible $\mathbf{c}^0 = \mathbf{1}$, the pose is frontal $\mathbf{h}^0 = \mathbf{0}$, and the there is no non-rigid deformation $\alpha^0 = \mathbf{0}$.
2. **for** *t=1, 2, ..., T or convergence* **do**
3.     Update the landmark visibility probabilities given the image, the landmark locations, and the head pose.;
4. 
$$f_t : I, \mathbf{x}^{t-1}, \mathbf{h}^{t-1} \to \Delta \mathbf{c}^t$$
$$\mathbf{c}^t = \mathbf{c}^{t-1} + \Delta \mathbf{c}^t$$
5.     Update the landmark locations given the image, the landmark locations, the landmark visibility probabilities, the head pose, and the non-rigid deformation.;
6. 
$$g_t : I, \mathbf{x}^{t-1}, \mathbf{c}^t, \mathbf{h}^{t-1}, \alpha^{t-1} \to \Delta \mathbf{x}^t$$
$$\mathbf{x}^t = \mathbf{x}^{t-1} + \Delta \mathbf{x}^t$$
7.     Update the head pose and non-rigid deformation given the landmark locations, the landmark visibility probabilities, and a pre-trained 3D deformable model.;
8. 
$$m_t : \mathbf{x}^t, \mathbf{c}^t, 3D\ model \to \mathbf{h}^t, \alpha^t$$
9. Output the estimated landmark locations $\mathbf{x}^T$, the predicted visibility probabilities $\mathbf{c}^T$, the estimated head pose $\mathbf{h}^T$, and the non-rigid deformation $\alpha^T$.

non-rigid deformation $\alpha^t$ simultaneously, we use the currently predicted landmark locations $\mathbf{x}^t$, the landmark visibility probabilities $\mathbf{c}^t$, and the 3D deformable model. Note that, in the third step for pose and deformation estimation, we follow model-based approach, and there is no learning involved. Therefore, we do not need any annotation for pose and deformation. We only need the landmark locations and occlusion labels, since learning is involved in the first two steps for landmark and occlusion predictions. Those three steps iterate until convergence. In the following, we discuss each step in details.

### 3.2. Predict the facial occlusion

In each cascade level, the first task is to estimate the facial occlusion. We need to update the landmark visibility probabilities $\mathbf{c}^t$ by predicting the landmark visibility probability updates $\Delta \mathbf{c}^t$. Intuitively, we can predict the visibility based on the local appearance information, which refers to the local image patches around the landmarks encoded with image features (e.g. SIFT features). The previously predicted head pose information is also an effective cue for estimation, since the occlusion could be caused by head poses. Overall, we use the linear regression model for the prediction.

$$\Delta \mathbf{c}^t = T_a^t \Phi(\mathbf{x}^{t-1}, I) + T_h^t \mathbf{h}^{t-1} \quad (2)$$
$$\mathbf{c}^t = \mathbf{c}^{t-1} + \Delta \mathbf{c}^t \quad (3)$$

In the first term, we predict the probability updates from the local facial appearance information, denoted as $\Phi(\mathbf{x}^{t-1}, I)$ for the image $I$ with regression parameters $T_a^t$. In the second term, we add the current pose angles as additional features for the prediction with linear regression parameters $T_h^t$. Note that, the prediction is bounded, since $\mathbf{c}^t$ represents probability vectors and they should be in the range $[0, 1]$.

The goal for model learning is to estimate the parameters of those linear regression models, including $T_a^t$, and $T_h^t$. We can get the ground truth landmark visibility probabilities ($c_d = 0$ for occluded point and 1 otherwise) given the ground truth facial landmark occlusion labels. Then we can calculate the ground truth landmark visibility probability updates $\Delta \mathbf{c}_i^{t,*} = \mathbf{c}_i^* - \mathbf{c}_i^{t-1}$, where $\mathbf{c}_i^*$ refers to the ground truth visibility probability (1 for visible points and 0 for occluded points). In addition, we have the previously estimated landmark locations $\mathbf{x}_i^{t-1}$, head pose angles $\mathbf{h}_i^{t-1}$, and deformation parameters $\alpha^{t-1}$. We can formulate parameter learning as a linear least squares problem with a closed form solution.

$$T_a^{t,*}, T_h^{t,*} = arg \min_{T_a^t, T_h^t} \sum_i \|\Delta \mathbf{c}_i^{t,*} - T_a^t \Phi(\mathbf{x}^{t-1}, I) - T_h^t \mathbf{h}^{t-1}\|^2 \quad (4)$$

### 3.3. Predict the facial landmark locations

In each cascade level, the second task is to update the facial landmark locations. We could use the local appearance information from the previously estimated landmark locations for the update, as in the traditional cascade regression framework [26]. However, due to the facial occlusion, the local appearance information may not be reliable. The head pose and the facial deformation can also be useful for landmark detection. To take all these factors into consideration, we predict the facial landmark location updates as follows:

$$\Delta \mathbf{x}^t = R_a^t [\sqrt{\mathbf{c}^t} \circ \Phi(\mathbf{x}^{t-1}, I)] + R_h^t \mathbf{h}^{t-1} + R_d^t \alpha^{t-1} \quad (5)$$
$$\mathbf{x}^t = \mathbf{x}^{t-1} + \Delta \mathbf{x}^t \quad (6)$$

In the first term, similar to [25], we propose to weigh the local appearance, denoted as $\Phi(\mathbf{x}^{t-1}, I)$, using the currently predicted landmark visibility probabilities $\mathbf{c}^t$, and use a linear regression model with parameter $R_a^t$ to predict the shape updates. "$\circ$" represents the point-wise product between the landmark occlusion probability of a particular point and its corresponding appearance information. The square root is used for better empirical performance. The intuition is that we want to weigh more on the appearance information from the visible points rather than that from the occluded points, since the occluded parts may not provide useful information for the landmark locations. In the second and third terms, we add the current pose and deformation information into the prediction with regression parameters $R_h^t$ and $R_d^t$, respectively.

For model learning, we can calculate the ground truth shape updates $\Delta \mathbf{x}_i^{t,*} = \mathbf{x}_i^* - \mathbf{x}_i^{t-1}$ for the arbitrary ith sample. Similar to learning the regression parameters for visibility prediction in section 3.2, we can formulate the parameter estimation problem in a weighted linear least squares formulation.

$$R_a^{t,*}, R_h^{t,*}, R_d^{t,*} = arg \min_{R_a^t, R_h^t, R_d^t} \sum_i \| \Delta \mathbf{x}_i^{t,*} \\ - R_a^t [\sqrt{\mathbf{c}^t} \circ \Phi(\mathbf{x}^{t-1}, I)] \\ - R_h^t \mathbf{h}^{t-1} - R_d^t \alpha^{t-1} \|_C^2 \quad (7)$$

In Equation 7, we introduce the diagonal matrix $C$ to handle missing facial landmark annotations. Due to the extreme head poses, some facial landmarks may not be visible for annotation. The corresponding element of $C$ is set to zero for completely occluded points, and one otherwise. Overall, the problem is a weighted linear least squares problem and it can be solved in a closed-form solution.

### 3.4. Predict the head pose and non-rigid deformation

In each cascade level, the third task is to predict the head pose angles $\mathbf{h}^t$ and facial deformation $\alpha^t$. We first build a 3D deformable model to capture the variations of frontal 3D facial shapes caused by the non-rigid facial deformation (e.g., cross-subject variations, expression variations). Given the 3D facial shapes as training data, we can learn the deformable model as shown in Equation 1 using the principal component analysis technique to generate the average 3D shape $\bar{s}$ and the orthonormal bases $B$.

Given the pre-trained 3D deformable model, the currently estimated facial landmark locations $\mathbf{x}^t$, and the facial visibility probabilities $\mathbf{c}^t$, we can simultaneously predict the head pose angles $\mathbf{h}^t$ and the facial deformation $\alpha^t$ by minimizing the projection error for all landmark points.

$$M^*, \alpha^* = \arg \min_{M, \alpha, t} \sum_k w_k(\|u_{d,k} - u_{p,k}\|^2 + \|v_{d,k} - v_{p,k}\|^2) \quad (8)$$

Here, $u_{d,k}$ and $v_{d,k}$ denote the column and row coordinates extracted from currently predicted 2D facial landmarks $\mathbf{x}^t$. $u_{p,k}$ and $v_{p,k}$ denote the column and row coordinates of the projected 2D landmarks based on the 3D deformable model with coefficients $\alpha^t$ and the current head pose $\mathbf{h}^t$ with weak-perspective projection.

$$\begin{bmatrix} u_k \\ v_k \end{bmatrix} = M \begin{bmatrix} x_k \\ y_k \\ z_k \end{bmatrix} + t \quad (9)$$

Here, $M$ is a 2 by 3 weak-perspective projection matrix that consists of two rows, representing respectively the scaled first and second rows of the rotation matrix. M hence encodes pose angles: pitch, yaw, and roll. $t$ presents 2D translation. As in section 3.3, we handle facial occlusion by introducing the weight $w_k$ for each point based on its visibility probability $c_k$. The intuition is that we weigh more on the projection errors produced by more visible landmark points. To solve Equation 8, we iteratively update the head pose angles and deformation parameters. While fixing one set of variables and estimating the other set of variables (e.g. M or $\alpha$), the problem becomes a weighted linear least squares problem with a closed-form solution.

## 4. Experimental results

### 4.1. Implementation details

#### 4.1.1 Model details

We use the SIFT [16] features to represent the local facial appearance information. We augment the training data by 8 random initializations of the landmarks in different scales and locations with consistent initializations of the pose and deformation for each sample. We trained the 3D deformable model using the 3D facial shape from BU4D-FE databases [27] provided in [22]. We retain 90% of the energy when choosing the number of principal components. The overall model contains four cascade iterations.

#### 4.1.2 Evaluation criteria

For facial landmark detection, we use two evaluation criteria. For images without self-occlusion (e.g. COFW [3]), we use the normalized error, which is defined as the distance between the predicted facial landmark locations and the ground truth landmark locations normalized by the inter-ocular distance (times 100%). We denote the error as normalized error. On images with self-occlusion (e.g. Multi-PIE [10]), since one eye may be occluded, we cannot use the normalized error. Therefore, we use the average absolute pixel distance. For facial occlusion prediction, we follow the previous works [3] and compare the recall values by fixing the precision value to 80%. For the evaluation of head pose estimation algorithms, we use the mean absolute error, which calculates the difference between the estimated angles and the ground truth angles. On databases with discrete head pose labels, we also include the head pose classification accuracy, which calculates the percentage of images that have an error less than 7.5 degrees.

## 4.2. Comparison of the proposed method to state-of-the-art works

### 4.2.1 Evaluation of head pose estimation on BU database

In the first experiment, we evaluate the proposed method on the Boston University (BU) head tracking database [15]. The BU database contains facial videos of subjects with different translation and rotation movements. Following the previous works [15] [20] [24], we use the 45 sequences of 5 subjects with uniform lighting conditions. Since the BU database only provides the continuous pitch yaw roll pose label, we cannot train the joint framework. Therefore, we only evaluate the head pose estimation on the BU database, where the 68 facial landmarks are generated using the pre-trained detector [26].

The experimental results are shown in Table 1. Since there is limited facial occlusion on the BU database, all the methods perform reasonably well and our method achieves the best overall head pose estimation accuracy compared to other model-based methods [21][15][20][24]. Our method is more effective than [21], [20] and [15], since they rely on a general rigid 3D head model or the cylindrical models. It also performs better than the flexible model in [24], even though they further assume that the camera intrinsic parameters are known. We also run [2] (code provided by the authors), and their pose estimation errors are also worse than ours (cannot reproduce their reported results).

Table 1. Comparison of the head pose estimation methods (mean absolute errors) on BU database.

| Method | Pitch | Yaw | Roll | Average |
|---|---|---|---|---|
| Rigid model [21] | 11.9 | 5.2 | 2.8 | 6.6 |
| Cylindrical [15] | 6.6 | 3.3 | 9.8 | 6.4 |
| Cylindrical+AAM [20] | 5.6 | 5.4 | 3.1 | 4.7 |
| Deformable model [24] | 4.3 | 6.2 | 3.2 | 4.6 |
| 3D CLM [2] | 6.0 | 3.9 | 3.7 | 4.5 |
| ours | 5.3 | 4.9 | 3.1 | 4.4 |

### 4.2.2 Evaluation of proposed joint method on COFW database with severe facial occlusion

In the second experiment, we evaluate the overall framework on the Caltech Occluded Faces in the Wild (COFW) database [3]. The COFW database contains facial images (Figure 2 (a)) with severe facial occlusions in arbitrary head poses, facial deformations collected from the website. There are 1345 training images and 507 testing images, with 29 facial landmark location annotations and their occlusion labels. Since COFW only provides the landmark annotation and occlusion labels, we can just evaluate the landmark detection and occlusion prediction accuracy.

The experimental results are shown in Table 2. We compared the proposed method with other state-of-the-art works, including the CRC [8], OC [9], RCPR [3], CRC [8], ESR [4], and FPLL [31], where the first three methods are specifically designed to handle facial occlusion. For both landmark detection and occlusion prediction, the proposed method outperforms all of the existing works, and its performance is closest to human performance. Our results on COFW outperforms the basic version of [25] (baseline in Tab. 1 of [25]), which demonstrates the importance of adding the pose/deformation information to help landmark/occlusion predictions. We are slightly worse than their full version with occlusion pattern and shape features. We can further include occlusion patterns and shape features to improve the performances of our method.

Table 2. Comparison of facial landmark detection errors (normalized errors w.r.t. inter-ocular distance) and occlusion prediction results on COFW database (29 points) [3].

| Method | Landmark error | Occlusion (precision/recall) |
|---|---|---|
| Human | 5.6 [3] | - |
| CRC [8] | 7.30 | - |
| OC [9] | 7.46 | 80.8/37.0% |
| RCPR [3] | 8.50 | 80/40% |
| ESR [4] | 11.20 | - |
| FPLL [31] | 14.40 | - |
| SDM [26] | 7.70 | - |
| ours | 6.40 | 80/44.43% |

### 4.2.3 Evaluation of proposed joint method on Multi-PIE database with varying head poses and self-occlusion

In the third experiment, we evaluate the full model on the MultiPIE database [10] for both head pose estimation and facial landmark detection. The MultiPIE database contains facial images (Figure 2 (b)) with varying illuminations, head poses, and facial deformations. There are 13 head poses with yaw angles ranging from -90 to 90 degrees with a 15-degree difference between two angles. The facial deformations are induced by different expressions, including happiness, surprise, etc. The facial landmark locations and head pose labels are provided. In the experiments, we use the facial images from the first 150 subjects for training and use the subjects with IDs between 151 and 200 as testing data. In the MultiPIE database, the facial occlusion is caused by self-occlusion due to extreme head poses.

Table 3. Facial landmark detection (pixel errors) and head pose estimation (mean absolute error, classification rate: error less than 7.5 degree) results on MultiPIE database (51 points) with the proposed method.

|  |  | 90 | 75 | 60 | 45 | 30 | 15 | 0 |
|---|---|---|---|---|---|---|---|---|
|  | landmark error | 3.41 | 3.33 | 3.17 | 4.16 | 2.96 | 3.19 | 3.37 |
| pose estimation | classification accuracy | 62.5 | 62.3 | 49.1 | 78.4 | 82.6 | 71.0 | 89.3 |
| pose estimation | yaw | 5.27 | 7.25 | 8.33 | 5.81 | 4.99 | 5.90 | 3.85 |
|  |  | -15 | -30 | -45 | -60 | -75 | -90 |  |
|  | landmark error | 3.39 | 3.09 | 3.99 | 3.13 | 4.03 | 4.89 |  |
| pose estimation | classification accuracy | 96.5 | 96.6 | 94.9 | 31.3 | 56.3 | 12.5 |  |
| pose estimation | yaw | 3.00 | 2.87 | 3.20 | 12.5 | 7.76 | 15.8 |  |

Table 4. Comparison of head pose estimation methods on MultiPIE database (mean absolute error, classification rate: error less than 7.5 degree).

|  | PCR [11] | linear PLS [11] | kPLS [11] | ours |
|---|---|---|---|---|
| yaw | 11.03 | 9.11 | 5.31 | 5.36 |
| classification accuracy | 48.33% | 57.22% | 79.48% | 77.1% |

Table 5. Comparison of landmark detection (average pixel errors) on MultiPIE database (51 points).

| near-frontal | | | | | | | all poses |
|---|---|---|---|---|---|---|---|
| CLM [18] | FPLL [31] | Pose-free [28] | Deep3D [30] | 3D CLM [2] | Chehra [1] | ours | ours |
| 4.75 | 4.39 | 7.34 | 5.74 | 5.30 | 4.09 | 3.51 | 3.50 |

Table 3 shows the performance of the proposed method on facial images with different yaw angles. For landmark detection, the average pixel errors are similar across different head poses. However, for head pose estimation, the errors increase on images with extreme head poses. This is due to the fact that for images with extreme head poses, the numbers of visible points decrease. With limited information, the prediction accuracy would decrease.

Comparisons of the proposed method to other state-of-the-art works are shown in Table 4 and Table 5. For head pose estimation, we compare the proposed method to the learning-based head pose estimation algorithms, including the Principal Component Regression (PCR), linear Partial Least Squares (PLS), and kernel PLS (kPLS) methods [11]. Our method is more accurate than PCR and linear PLS, and it is comparable to kPLS. The model-based methods [2] that take all landmarks may fail on the images with large head poses.

For landmark detection, our method outperforms CLM [18], FPLL [31], Pose-free [28], Deep 3D [30], 3D CLM [2], and Chehra [1] on near-frontal facial images. The performances of the proposed method on images of all 13 poses are similar to that of near-frontal images.

### 4.3. Further evaluation of the proposed method

#### 4.3.1 Effectiveness of the interaction among landmark, pose and deformable estimation

One major benefit of the proposed method is to leverage the interactions among landmarks, pose, and deformation to boost the performances of all tasks. In this section, we show empirical study about how the joint interactions would improve the performances.

First, we show how the occlusion, pose and deformation estimation will improve facial landmark detection on the COFW database in Table 6. The baseline method is a conventional cascade regression method for landmark detection [26], and it doesn't consider the facial occlusion, head pose, and facial deformation. Our method can add the occlusion prediction (ours_occlusion) without the head pose and deformation estimation to help in facial landmark detection. We can also additionally add the head pose and facial deformation (second term in Equation 2, second and third terms in Equation 5) to have the full model (ours_all) for the prediction. From the comparison, we see that by adding the occlusion (ours_occlusion), the landmark de-tection performance is better than the baseline. By further adding the head pose and deformation, the full model (ours_all) achieves the best performance. The experiments demonstrate the effectiveness of utilizing the interactions among landmarks, pose, and deformation and the occlusion prediction.

Table 6. Effectiveness of joint occlusion, pose, and deformation estimation for landmark detection on COFW database (normalized errors w.r.t inter-ocular distance, 29 points).

| Method | baseline [26] | ours_occlusion | ours_all |
|---|---|---|---|
| error | 7.70 | 6.61 | 6.40 |

Second, we illustrate why the occlusion prediction is important for head pose and deformation estimation on the MultiPIE database. As shown in Figure 4, due to self-occlusion, it's difficult to detect the facial landmarks on the occluded facial part. Without considering the landmark occlusion, the pose estimation algorithms may take all the facial landmarks into consideration and lead to incorrect pose estimation results. Our method jointly performs landmark detection and occlusion estimation. Therefore, the method only uses the visible points for pose estimation. For example, by fitting all points for Figure 4 (a), the estimated yaw angle is about 26 degree, which significantly differs from the ground truth (yaw angle is 90 degree). If we consider the occlusion label, we can use the visible points and accurately estimate the yaw angle as 90 degree.

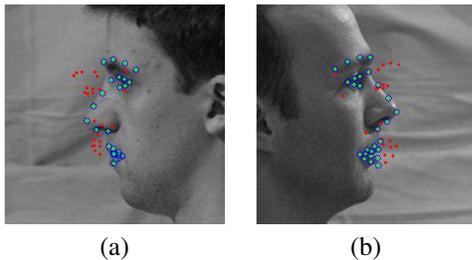

Figure 4. Facial landmark detection and occlusion prediction results on sample images. Green points: visible landmarks. Red points: occluded landmarks.

#### 4.3.2 Convergence study

Since the proposed method is an iterative cascade method, we need to study its convergence property. Figure 5 and 6 show the performances of the proposed method across different iterations on COFW and MultiPIE databases, respectively. For COFW, we show landmark detection and occlusion prediction performances. For MultiPIE, we show landmark detection and pose estimation accuracies. As can be seen, the proposed method converges quickly for landmark detection, occlusion prediction and pose estimation.

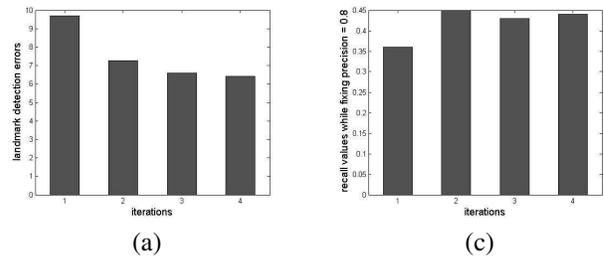

Figure 5. Performance of the proposed method across iterations on COFW database. (a) Landmark detection errors (normalized errors). (b) Occlusion prediction accuracy (recall while fixing precision = 0.8).

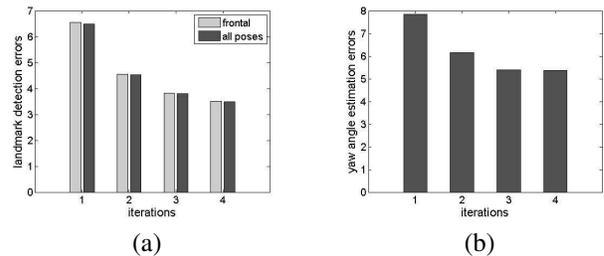

Figure 6. Performance of the proposed method across iterations on MultiPIE database. (a) Landmark detection errors (pixel errors). (b) Head pose estimation errors for the yaw angle (mean absolute error).

## 5. Conclusion

In this work, we propose a unified framework that can simultaneously perform facial landmark detection, head pose estimation and deformation estimation under facial occlusion. With a cascade iterative procedure augmented with model-based pose estimation, we iteratively predict the facial occlusion, facial landmark locations, head pose angles and facial deformation. The iterative cascade procedure allows us to fully exploit their joint relationships. The experiments demonstrate the effectiveness of the proposed methods on benchmark databases compared to state-of-the-art works for landmark detection, occlusion predictions and head pose estimations.

In the future, we would further evaluate the framework on more "in-the-wild" databases with the joint landmark, poses, and deformation annotations. In addition, although the proposed method solves a specific vision problem, it demonstrates the power of leveraging the relationships among related tasks with the cascade iterative procedure. It can be applied to other problems that involve multiple tasks, such as joint object detection, image segmentation, and scene understanding.

**Acknowledgements:** The work described in this paper was partly supported by IBM Ph.D. fellowship and by the RPI-IBM Cognitive and Immersive Systems Laboratory (CISL) project.